\def\year{2020}\relax
\documentclass[letterpaper]{article} 
\usepackage{aaai20}  
\usepackage{times}  
\usepackage{helvet} 
\usepackage{courier}  
\usepackage[hyphens]{url}  
\usepackage{graphicx} 
\usepackage{graphicx}  
\usepackage{amsmath}
\usepackage{amssymb}

\title{Every Frame Counts: Joint Learning of Video Segmentation and Optical Flow}

\author{
  Mingyu Ding\textsuperscript{\rm 1,3}~~ Zhe Wang\textsuperscript{\rm 5}~~ Bolei Zhou\textsuperscript{\rm 4}~~ Jianping Shi\textsuperscript{\rm 5}~~ Zhiwu Lu\textsuperscript{\rm 1,2}\thanks{Corresponding author.}~~ Ping Luo\textsuperscript{\rm 3}\\
 \textsuperscript{\rm 1}Gaoling School of Artificial Intelligence, Renmin University of China, Beijing 100872, China\\
 \textsuperscript{\rm 2}Beijing Key Laboratory of Big Data Management and Analysis Methods, Beijing 100872, China\\
 \textsuperscript{\rm 3}The University of Hong Kong~~  \textsuperscript{\rm 4}The Chinese University of Hong Kong~~ \textsuperscript{\rm 5}SenseTime Research\\
  \texttt{dingmyu@gmail.com}~~~~~~
  \texttt{luzhiwu@ruc.edu.cn}~~~~~~~
  \texttt{pluo@cs.hku.hk}
  \\
}

\begin{document}

\maketitle

\begin{abstract}
A major challenge for video semantic segmentation is the lack of labeled data. In most benchmark datasets, only one frame of a video clip is annotated, which makes most supervised methods fail to utilize information from the rest of the frames. To exploit the spatio-temporal information in videos, many previous works use pre-computed optical flows, which encode the temporal consistency to improve the video segmentation. However, the video segmentation and optical flow estimation are still considered as two separate tasks. In this paper, we propose a novel framework for joint video semantic segmentation and optical flow estimation. Semantic segmentation brings semantic information to handle occlusion for more robust optical flow estimation, while the non-occluded optical flow provides accurate pixel-level temporal correspondences to guarantee the temporal consistency of the segmentation. Moreover, our framework is able to utilize both labeled and unlabeled frames in the video through joint training, while no additional calculation is required in inference. Extensive experiments show that the proposed model makes the video semantic segmentation and optical flow estimation benefit from each other and outperforms existing methods under the same settings in both tasks.
\end{abstract}

\section{Introduction}
Video semantic segmentation, as an important research topic for applications such as robotics and autonomous driving, still remains largely unsolved.
Current video segmentation methods mainly face two aspects of challenges: inefficiency and lack of labeled data. On the one hand, since frame-by-frame labeling of the video is time consuming, the existing data set contains only one annotated frame in each snippet, thus making the problem more challenging. On the other hand, to incorporate temporal information of the video, existing methods deploy feature aggregation modules to improve the segmentation accuracy, which leads to inefficiency during the inference phase.

\begin{figure}[t]
\centering
\includegraphics[width=0.99\columnwidth]{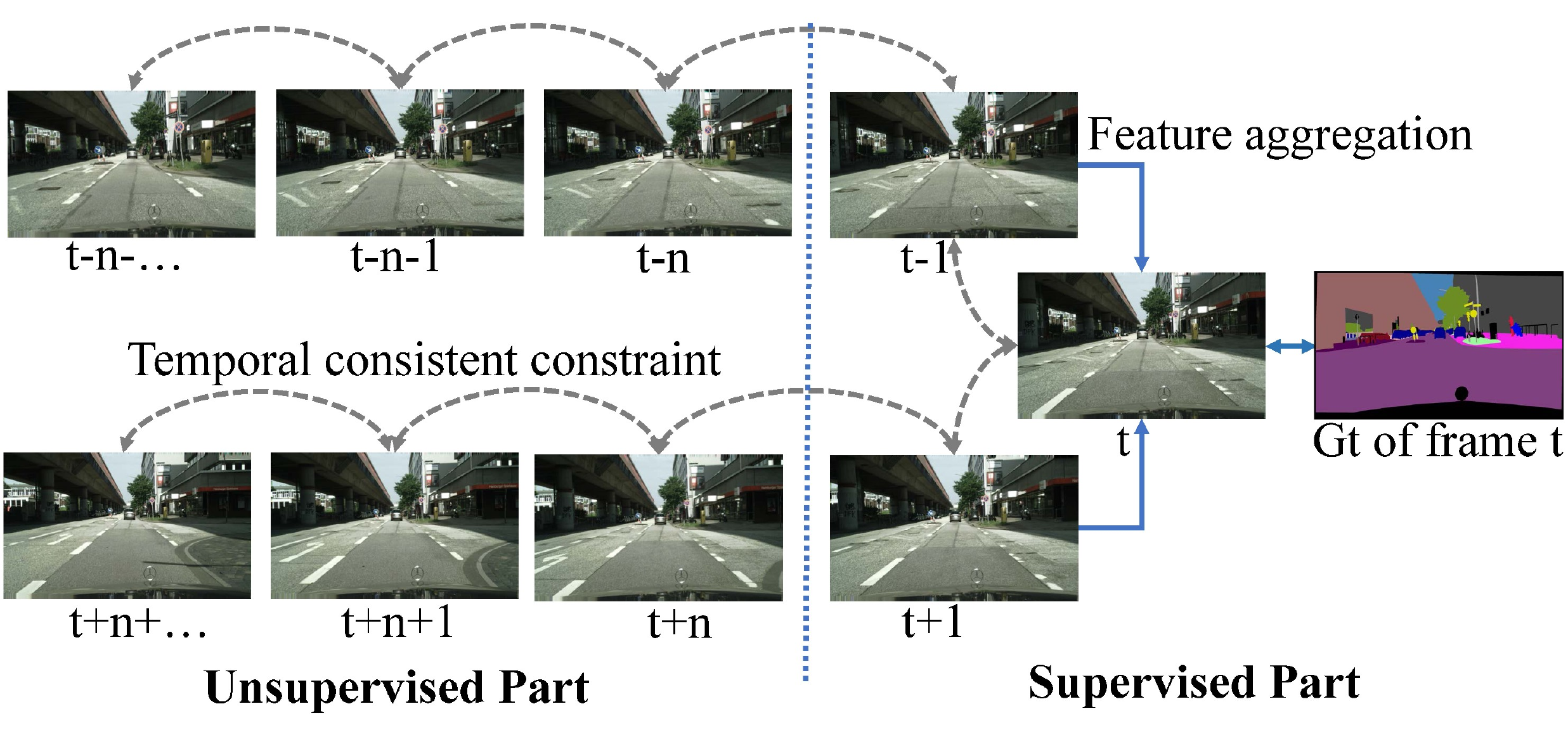}
\caption{
Illustration of a video snippet that contains 30 frames with only one annotated frame t. Unlike previous models that only utilize those frames close to the frame with ground-truth by feature aggregation (solid lines), our model makes full use of all frames in the video with temporal consistent constraints (dashed lines).
}
\label{fig:data}
\end{figure}

Optical flow, which encodes the temporal consistency across frames in video, has been used to improve the segmentation accuracy or speed up the segmentation computation.
For examples, the methods \cite{li2018low,zhu2017deep,shelhamer2016clockwork} reuse the features in previous frames to accelerate computation. However, doing so will result in a decrease in the accuracy of the segmentation, and such methods are not considered in this paper. On the other hand, the methods \cite{fayyaz2016stfcn,jin2017video,gadde2017semantic,nilsson2016semantic,hur2016joint} model multiple frames by flow-guided feature aggregation or a sequence module for better segmentation performance, which increases computational cost.
Our motivation is to use optical flow to exploit temporal consistency in the semantic feature space for training better models, with no cost in inference time.

Current video segmentation datasets such as \cite{cordts2016cityscapes} only annotate a small fraction of frames in videos. Existing methods focus on combining features of consecutive frames to achieve better segmentation performance. These methods can only use a small portion of frames in the video. Moreover, additional data is needed for training the feature aggregation module (FlowNet) in flow-guided methods \cite{nilsson2016semantic}.

To address the two challenges of video semantic segmentation, we propose a joint framework for semantic segmentation and optical flow estimation to fully utilize the unlabeled video data and overcome the problem of pre-computing optical flow.
Semantic segmentation introduces semantic information that helps identify occlusion for more robust optical flow estimation. Meanwhile, non-occluded optical flow provides accurate pixel-level  correspondences to guarantee the temporal consistency of the segmentation. These two tasks are related through temporal and spatial consistency in the designed network.
Therefore, our model benefits from learning all the frames in the video without feature aggregation, which means that there is no extra calculation in inference. To the best of our knowledge, this is the first framework that joint learns these two tasks in an end-to-end manner.

We summarize our contributions as follow: (1) We design a novel framework for joint learning of video semantic segmentation and optical flow estimation with no extra calculation in inference. All the video frames can be used for training with the proposed temporally consistent constraints.
(2) We design novel loss functions that handle flow occlusion in both two tasks, which improves the training robustness.
(3) Our model makes the video semantic segmentation and optical flow estimation mutually beneficial and is superior to existing methods under the same setting in both tasks.

\section{Related Work}

\textbf{Video Segmentation.} Video semantic segmentation considers temporal consistency of consecutive frames compared to semantic segmentation. Existing methods mainly fall into two categories. The first category aims to accelerate computation by reusing the features in previous frames. Shelhamer \textit{et al.} proposed a Clockwork network \cite{shelhamer2016clockwork} that adapts multi-stage FCN and directly reuses the second or third stage features of preceding frames to save computation. \cite{zhu2017deep} presented the Deep Feature Flow that propagates the high level feature from the key frame to current frame by optical flow learned in FlowNet \cite{dosovitskiy2015flownet}. \cite{li2018low} proposed a network using spatially variant convolution to propagate features adaptively and an adaptive scheduler to ensure low latency. However, doing so will result in a decrease of accuracy, which is not considered in this paper.

Another category focuses on improving accuracy of segmentation by flow-guided feature aggregation or some sequence module. Our model falls into this category. \cite{fayyaz2016stfcn} proposed to combine the CNN features of consecutive frames through a spatial-temporal LSTM module. \cite{gadde2017semantic} proposed a NetWarp module to combine the features wrapped from previous frames with flows and those from the current frame to predict the segmentation. \cite{nilsson2016semantic} proposed gated recurrent units to propagate semantic labels. \cite{jin2017video} proposed to learn from unlabeled video data in an unsupervised way through a predictive feature learning model (PEARL). However, such methods require additional feature aggregation modules, such as flow warping modules and sequence modules, which greatly increase the computational costs during the inference phase. Moreover, the feature aggregation modules of these methods can only process the annotated frame and several frames around it, while the rest of the frames are largely discarded in the training. In contrast, our method has two parallel branches for semantic segmentation and optical flow estimation, which reinforce each other in training but adds no extra calculation in inference. Furthermore, we can also leverage all video frames to train our model, with our temporally consistent constraint.

There are also other video segmentation methods with different settings.
\cite{kundu2016feature} applied a dense random field over an optimized feature space for video segmentation. \cite{chandra2018deep} introduced densely-connected spatio-temporal graph on deep Gaussian Conditional Random Fields. \cite{hur2016joint} estimates optical flow and temporally consistent semantic segmentation based on an 8-DoF piecewise-parametric model with a superpixelization of the scene. However, the iterative method based on superpixel cannot benefit from unsupervised data nor be optimized end-to-end. Our model can benefit from unsupervised data and be trained in an end-to-end deep manner, making the two tasks mutually beneficial. \cite{cheng2017segflow} proposed to learn video object segmentation and optical flow in a multi-task framework, which focuses on segmenting instance level object masks. Both optical flow and object segmentation is learned in a supervised manner.
In comparison, our task is semantic segmentation for the entire image and our optical flow is learned unsupervisedly. The two tasks cannot be directly compared.

\begin{figure*}[t]
\centering
\includegraphics[width=1.8\columnwidth]{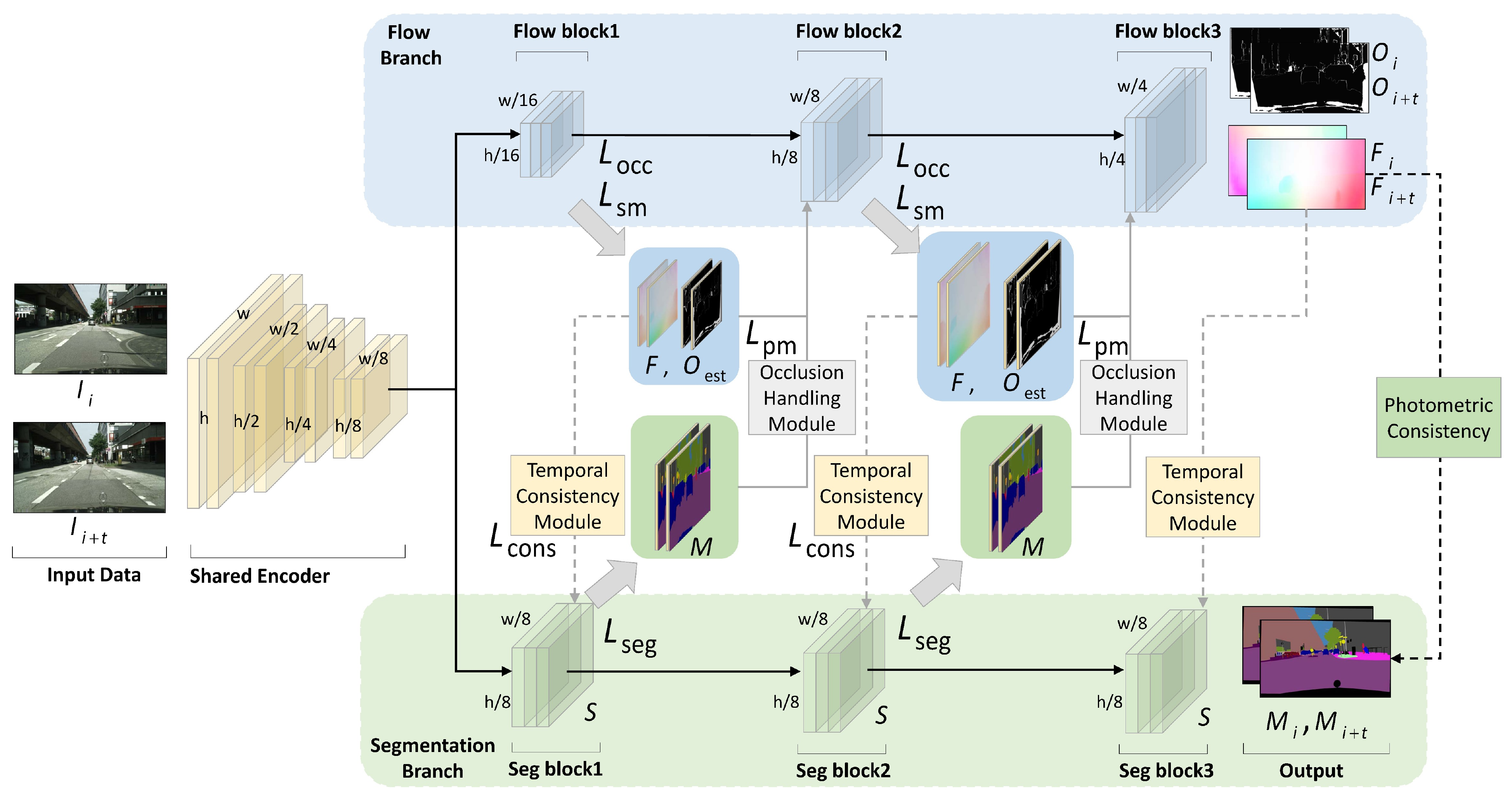}
\caption{The overall pipeline of our joint learning framework. The blocks represent the feature maps of our model, the gray dashed line represents the temporally consistent constraints. The gray solid line represents the occlusion handling module with the inconsistency of the segmentation maps.}
\label{fig:overview}
\end{figure*}

\textbf{Optical Flow Estimation} Optical flow estimation requires finding correspondences between two input images. FlowNet and FlowNet2.0 \cite{dosovitskiy2015flownet,ilg2017flownet} directly compute dense flow prediction on every pixel through fully convolutional neural networks. PWC-Net \cite{sun2018pwc} uses the current optical flow estimate to warp the CNN features of the second image. \cite{patraucean2015spatio} introduced a spatio-termporal video autoencoder based on an end-to-end architecture that allows unsupervised training for motion prediction. \cite{jason2016back,meister2017unflow,ren2017unsupervised} utilizes the Spatial Transformer Networks \cite{jaderberg2015spatial} to warp current images and measures photometric constancy. \cite{wang2018occlusion,janai2018unsupervised} models occlusion explicitly during the unsupervised learning of optical flow. In this work, the occlusion mask is refined by introducing the semantic information in our proposed approach. Moreover, the unsupervised optical flow estimation framework can be further extended to estimate monocular depth, optical flow and ego-motion simultaneously in an end-to-end manner \cite{yin2018geonet}.  \cite{ren2017cascaded} proposed a cascaded classification framework that accurately models 3D scenes by iteratively refining semantic segmentation masks, stereo correspondences, 3D rigid motion estimates, and optical flow fields.

\section{Methodology}

Our framework, EFC model (Every Frame Counts), learns video semantic segmentation and optical flow estimation simultaneously in an end-to-end manner. In the following, we first give an overview of our framework and then describe each of its components in detail.

\subsection{Framework Overview}

An overview of our EFC model is illustrated in Figure~\ref{fig:overview}.
The input to our model is a pair of images $I_i, I_{i+t}$, randomly selected from near-by video frames with $t \in [1, 5]$.
If either $I_i$ or $I_{i+t}$ has semantic labels, we can update weights of the network by supervised constraints with semantic labels as well as unsupervised constraints from near-by frame correspondence.
It propagates semantic information across frames, and jointly optimize the semantic component and optical flow component to reinforce each other.
Otherwise, only unsupervised consistency information can be utilized, and our network can benefit from the improvement in the optical flow component.

Specifically, our network consists of the following three parts, \textit{i}.\textit{e}., the shared encoder part, the segmentation decoder part and the flow decoder part. The shared encoder contains layers 1-3 of ResNet \cite{he2016deep}. It is helpful since semantic and flow information exchange among the representation, increasing the representation ability compared to \cite{zhao2017pyramid}. The semantic decoder is adopted from layer 4 of ResNet if semantic label exists. The flow decoder combines intermediate feature from frame $I_i$ and $I_{i+t}$ via a correlation layer following \cite{ilg2017flownet} to predict optical flow. A smoothness loss on flow result is applied to improve flow quality.

To enable end-to-end cross frame training without optical flow label, we design a temporal consistency module. It can warp both input image pairs and intermediate feature pairs via the predicted flow and regresses warping error as the photometric loss and temporal consistency loss accordingly. To further increase robustness with heavy occlusion, where the predicted optical flow is invalid, we introduce the occlusion handling module with an occlusion aware loss. The occlusion mask is also learned end-to-end and improves with better predicted optical flow.
In the following, we will introduce each module of our model in detail.

\begin{figure*}[t]
\centering
\includegraphics[width=1.8\columnwidth]{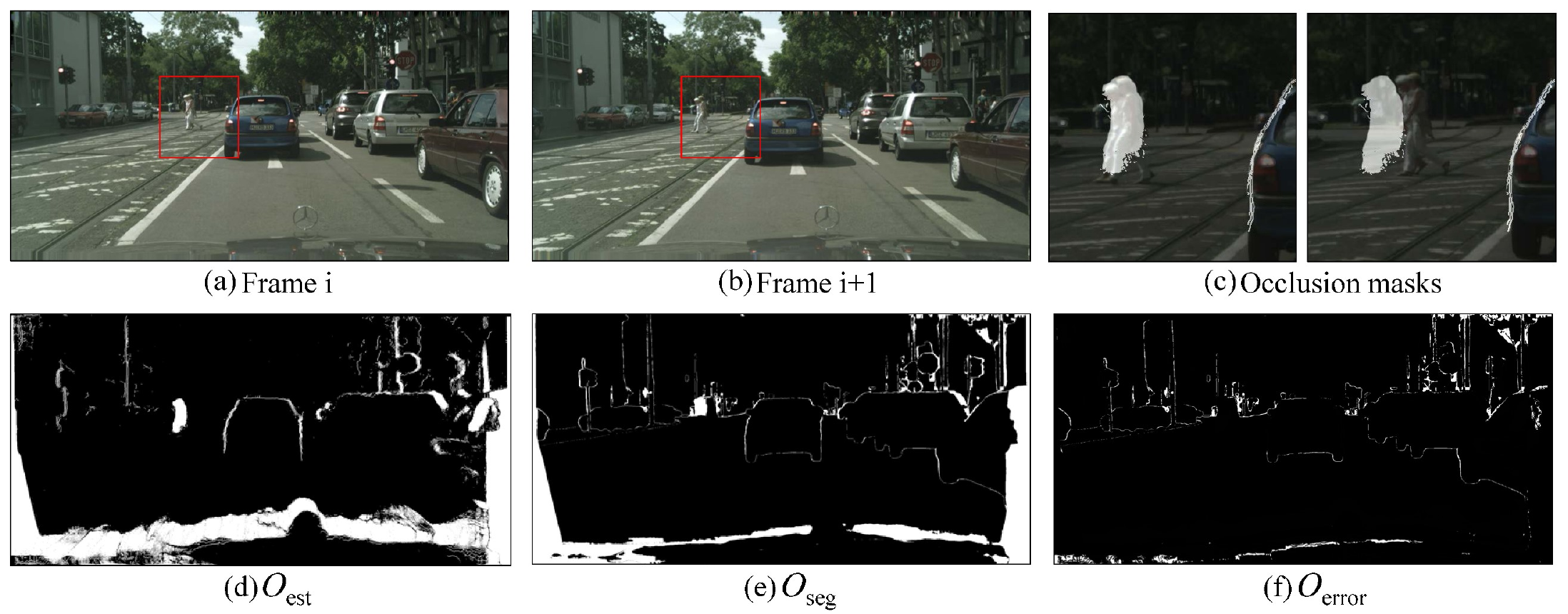}
\caption{Two types of occlusion masks are applied in our model. $(c)$ shows the zoom-in occlusion masks inside the red rectangle region of $(a)$. $(d)$ is the occlusion mask $O_\text{est}$ which is estimated by the non-occluded flow branch. $(e)$ is the occlusion mask $O_\text{seg}$ obtained from the inconsistency of the segmentation maps. The error mask $O_\text{error}$ is shown in $(f)$.}
\label{fig:mask}
\end{figure*}

\subsection{Temporally Consistent Constraint}

Photometric consistency is usually adopted in optical flow estimation, where the first frame is warped to the next by optical flow and the warping loss can be used for training the network. In this work, we generalize the photometric loss to the feature domain. As the convolution neural network is translation invariant, the feature maps of adjacent frames should also follow the temporally consistent constraint.

More specifically, for a pair of video frames  $I_i$ and $I_{i+t}$, we feed them into the shared encoder network to extract their feature maps $S_i$ and $S_{i+t}$. Since we learn both forward and reverse optical flows $F_{i\mapsto i+t}, F_{i+t \mapsto i}$ simultaneously, we then warp  $S_{i+t}, S_i$  to $ S_{i}', S_{i+t}'$ by flow $F_{i\mapsto i+t}, F_{i+t \mapsto i}$ so that $S'$ is expected to be consistent with feature map $S$. Formally, $S_{i}'$ can be obtained by
\begin{equation}
\begin{small}
S_{i}' = \text{Warp}(\mathcal{S}_{i+t}, F_{i\mapsto i+t}),
\label{eq:Warp}
\end{small}
\end{equation}
where we adopt the differentiable bilinear interpolation for warping. Note that the warping direction is different from the flow direction. However, the flow can be invalid in occluded regions. So we estimate the occlusion maps $O_\text{est}^i$ and $O_\text{est}^{i+t}$ by checking if one pixel has a corresponding pixel in the adjacent frame.  With the occlusion maps, we avoid penalizing the pixels in the occluded regions. The temporal consistency loss is thus defined as:
\begin{align}
\begin{small}
L_\text{cons} = \sum_{x,y}{(1-O_\text{est}^{xy}) \cdot \| S'^{xy} - S^{xy} \|^2},
\label{eq:loss_Consistency}
\end{small}
\end{align}
where $S^{xy}$ is the feature at location $(x,y)$. Notice that we take warping constraints in both directions for training.

The temporal consistency loss introduces a temporal regularization on the feature space, thus allowing our model to be trained with unlabeled video data. When the label is unavailable, our model can still benefit from the temporal consistency constraint.

\subsection{Occlusion Estimation}

Our model learns occlusion in a self-supervised manner. The occlusion defined here is a general term. By occlusion we refer to the pixels that are photometric inconsistent in two given frames, which can be caused by real occlusion by objects, in-and-out of image, change of view angle or so.
The occlusion and the optical flow estimation network share most of the parameters. For each block in non-occluded flow branch, we add two convolutional layers with very few channels and a sigmoid layer for occlusion estimation. By backward optical flow $F_{i+t\mapsto i}$, we can calculate the correspondence between the two frames $I_i, I_{i+t}$ in pixel-level. We decompose optical flow into vertical  $F_{i+t\mapsto i}(y,x,1)$ part and horizontal  $F_{i+t\mapsto i}(y,x,0)$ part. Then we have:
\begin{align}
& y_{i+t} = y_{i}-F_{i+t\mapsto i}(y_{i+t},x_{i+t},1), \notag \\
& x_{i+t} = x_{i}-F_{i+t\mapsto i}(y_{i+t},x_{i+t},0).
\end{align}

The occlusion mask $\hat{O}_i$ for the backward flow $F_{i+t\mapsto i}$ can be formulated as: $\hat{O}_i (y_i,x_i) = 0$ if there is a corresponding pixel $(y_{i+t},x_{i+t})$ in $I_{i+t}$ ($0\leq x_{i+t} < w$ \& $0\leq y_{i+t} < h$), otherwise $\hat{O}_i (y_i,x_i) = 1$. Then cross entropy with a penalty is used for occlusion estimation. The network mimics $\hat{O}$, and produces finer masks by our loss function $L_\text{occ}$:
\begin{align}
\begin{small}
L_\text{occ} = -\sum_{x,y}\log p(O_\text{est}^{xy} = \hat{O}^{xy} ) - \alpha e^{-O_\text{est}^{xy}}.
\label{eq:loss_Occlusion}
\end{small}
\end{align}
Since we do not calculate the consistency loss of the occlusion region, the network tends to predict more occlusion regions. So the second penalty term is used to prevent excessive occlusion prediction. The larger $\alpha$ is, the greater penalty for the occlusion region, and the smaller the occlusion region predicted. We tried different $\alpha$ values between 0 and 1, and found that 0.2 is the best.

\subsection{Optical Flow Estimation}

Similar to \cite{yin2018geonet,jason2016back,wang2018occlusion}, optical flow can be learned in a self-supervised manner. More specifically, the first frame can be warped to the next frame by the predicted optical flow, and the photometric consistency and motion smoothness are exploited for training. Photometric consistency is to reconstruct the scene structure between two frames and motion smoothness is to filter out erroneous predictions and preserve sharp details. In this work, we observe that semantic information can be leveraged by joint training to help estimation of optical flow.

As shown in Figure~\ref{fig:overview}, the semantic maps $M$ introduce semantic information on the likely physical motion of the associated pixels. Besides, we generate error masks which point out the inaccurate regions of the optical flow for robust optical flow estimation. As illustrated in Figure~\ref{fig:mask}, we first calculate an inconsistent mask $O_\text{seg} = (M \neq M')$ between our two branches, where $M'$ is the warped segmentation prediction with bilinear interpolation. Then we define the error mask $O_\text{error}$ as:
\begin{equation}
\begin{small}
O_\text{error} = \mathop{\max} (O_\text{seg} - O_\text{est}, 0).
\label{eq:mask_error}
\end{small}
\end{equation}

The inconsistent mask of two segmentation maps should contain the occlusion mask and the offset due to in-accurate optical flow. To unify these two masks, we simply double the weight of the error mask region and ignore the occlusion mask region during optical flow learning. Our photometric loss $L_{pm}$ can be calculated with the following equation:
\begin{align}
& L_\text{pm} = \sum_{x,y}(\mathcal{G}(I,I')^{xy} \cdot (1+O_\text{error}^{xy}-O_\text{est}^{xy})),\nonumber\\
& \mathcal{G}(I,I') =
\beta\frac{1-SSIM(I,I')}{2}+(1-\beta) \|I - I'\|_1,
\label{eq:loss_Photometric}
\end{align}
where $I'$ is a warped image, $SSIM$ is the per pixel structural similarity index measurement \cite{wang2004image}, $\mathcal{G}$ denotes the loss map, which indicates the weight to penalize at different locations. Here we adopt a linear combination of two common metrics for estimating similarity of the original image and the warped one. Intuitively, the pixels perfectly matched indicate the estimated flow is correct and get less penalized in the photometric loss. $\beta$ is taken to be 0.85 as in \cite{yin2018geonet}. Following \cite{jason2016back,yin2018geonet},  The smoothness loss is defined as:
\begin{align}
\begin{small}
L_\text{sm} = \sum_{x,y}\mid\Delta F(x,y) \mid \cdot (e^{-\mid\Delta I(x,y) \mid}),
\label{eq:loss_Smooth}
\end{small}
\end{align}
where $\Delta$ is the vector differential operator. Note that both the photometric and smoothness losses are calculated on multi-scale blocks and two directions.

\subsection{Joint Learning}

For the frames that have ground truths $M_\text{gt}$, we use the standard log-likelihood loss for semantic segmentation:
\begin{equation}
\begin{small}
L_\text{seg} = -\sum_{x,y}\log p(M^{xy} = M_\text{gt}^{xy} ).
\label{eq:loss_Seg}
\end{small}
\end{equation}
To summarize, our final loss for the entire framework is:
\begin{align}
\begin{small}
L =  L_\text{seg} + \lambda_\text{cons} L_\text{cons} + \lambda_\text{occ}L_\text{occ} + \lambda_\text{sm}L_\text{sm} + L_\text{pm},
\label{eq:loss_All}
\end{small}
\end{align}
where $\lambda_\text{cons}$, $\lambda_\text{occ}$, and $\lambda_\text{sm}$ denote the weights for multiple losses. Our entire framework is thus trained end-to-end.

\section{Experiments}

\subsection{Dataset and Setting}

\textbf{Datasets} We evaluate our framework for video semantic segmentation on the Cityscapes \cite{cordts2016cityscapes} and CamVid datasets \cite{brostow2009semantic}. We also report our competitive results for optical flow estimation on the KITTI dataset \cite{geiger2012we}.

Cityscapes \cite{cordts2016cityscapes} contains 5,000 sparsely labeled snippets collected from 50 cities in different seasons, which are divided into sets with numbers 2,975, 500, and 1,525 for training, validation and testing. Each snippet contains 30 frames, and only the 20th frame is finely annotated in pixel-level. 20,000 coarsely annotated images are also provided.

CamVid \cite{brostow2009semantic} is the first collection of videos with object class semantic labels, it contains 701 color images with annotations of 11 semantic classes. We follow the same split in \cite{kundu2016feature,nilsson2016semantic} with 367 training images, 100 validation images and 233 test images.

KITTI \cite{geiger2012we} is a real-world computer vision benchmark dataset with multiple tasks. The training data we use here is similar to \cite{yin2018geonet}, where the official training images are adopted as testing set. All the related images in the 28 scenes covered by testing data are excluded. Since there are no segmentation labels on our training set, we generate some coarse segmentation results as the segmentation ground truths through a model trained on Cityscapes.

\textbf{Evaluation Metrics} We report mean Intersection-over-Union (mIoU) scores for semantic segmentation task on Cityscapes and CamVid datasets. The optical flow performance for the KITTI dataset is measured by the average end-point-error (EPE) score.

\textbf{Implementation Details} Our framework is not limited to specific CNN architectures. In our experiments, we use the original PSPNet \cite{zhao2017pyramid} and the modified FlowNetS \cite{dosovitskiy2015flownet} as the baseline network unless otherwise specified. The FlowNetS is modified as follows: (1) share the encoder with PSPNet. (2) add two $3 \times 3$ convolution layers for occlusion estimation with 32 and 1 channels, respectively.
The loss weights are set to be $\lambda_{cons}=10, \lambda_{occ}=0.4 \ \textup{and}\  \lambda_{sm}=0.5$ for all experiments.

During training, we randomly choose ten pairs of images with $\Delta t \in [1, 5]$ from one snippet, five of which contain images with ground truths.
The training images are randomly cropped to $713 \times 713$. We also perform random scaling, rotation, flip and other color augmentations for data augmentation.
The network is optimized by SGD, where momentum and weight decay are set to 0.9 and 0.0001 respectively.
We take a mini-batch size of 16 on 16 TITAN Xp GPUs with synchronous Batch Normalization. We use the `poly' learning rate policy and set base learning rate to 0.01 and power to 0.9, as in \cite{zhao2017pyramid}. The iteration number for training process is set to 120K.

\begin{table}[t]
\caption{Ablation study for video semantic segmentation on the Cityscapes validation set. ResNet50-based PSPNet (single scale testing) is used as a baseline model.}
\begin{center}
\label{tab:ablation}
\tabcolsep0.7cm
\begin{tabular}{l|c}
\hline
Model & IoU ($\%$)\\
\hline
ResNet50 + PSPNet  &  76.20\\
+ $\textup{TCC}_{fix}$ &  77.02 \\
+ $\textup{TCC}_{single}$ &  77.58  \\
+ $\textup{TCC}_{single}$ + OM &  77.79  \\
+ $\textup{TCC}_{multi}$ + OM &  78.07  \\
+ $\textup{TCC}_{multi}$ + OM + UD & \textbf{78.44} \\
\hline
\end{tabular}
\end{center}
\end{table}

\subsection{Ablation Study}

To further evaluate the effectiveness of the proposed components, i.e., the joint learning, the temporally consistent constraint, the occlusion masks, and the unlabeled data, we conduct ablation studies on both the segmentation and optical flow tasks. All experiments use the same training setting.

For video segmentation, we make comparisons to five simplified versions on the Cityscapes validation set:
(1) $\textup{TCC}_{fix}$ -- temporally consistent constraint on a single pair of images with the fixed pre-trained FlowNetS.
(2) $\textup{TCC}_{single}$ -- temporally consistent constraint without the occlusion mask on a single pair of images.
(3) $\textup{TCC}_{single}$ + OM -- temporally consistent constraint with the occlusion mask on a single pair of images.
(4) $\textup{TCC}_{multi}$ + OM -- temporally consistent constraint with the occlusion mask on randomly selected five pairs of images.
(5) $\textup{TCC}_{multi}$ + OM + UD -- our full EFC model with unlabeled data.

\begin{table}[t]
\caption{Ablation study for optical flow estimation on the KITTI Dataset.}
\begin{center}
\tabcolsep0.7cm
\begin{tabular}{c||c|c}
\hline Method  & Noc & All \\ \hline\hline
UL  & 7.53 & 11.03 \\
UL + OE  & 7.23 & 8.72 \\
UL + TC  & 4.94 & 8.84 \\
UL + OE + TC  & 4.51 & 7.79 \\
EFC\_full  & \textbf{3.93} & \textbf{7.05} \\
\hline
\end{tabular}
\end{center}
\label{tab:kitti_ablation}
\end{table}

\begin{table}[t]
\caption{Comparative results of video segmentation on the Cityscapes test set. Notation: `PSP' -- the PSPNet trained with only finely annotated data, `PSP\_CRS' -- the PSPNet trained with both finely and coarsely annotated data, `C' -- whether coarsely annotated data is used, `IoU cls' -- average class IoU ($\%$), `IoU cat' -- average category IoU ($\%$).}
\tabcolsep0.15cm
\begin{center}
\begin{tabular}{c||c|c|c}
\hline Method & C & IoU cls & IoU cat  \\ \hline\hline
Clockwork \shortcite{shelhamer2016clockwork} &  & 66.4 & 88.6 \\ 
PEARL \cite{jin2017video} &  & 75.4 & 89.2 \\
LLVSS \cite{li2018low} &  & 76.8 & 89.8 \\
Accel \shortcite{jain2019accel} &  & 75.5 & -- \\
DFANet \cite{li2019dfanet} &  & 71.3 & -- \\
\hline
\hline
Dilation10 \shortcite{yu2015multi} & & 67.1 & 86.5 \\
Dilation10 + GRFP \shortcite{nilsson2016semantic} & & 67.8 & 86.7 \\
Dilation10 + EFC (Ours) &  & \textbf{68.7} & \textbf{87.3} \\
\hline
\hline
PSP \cite{zhao2017pyramid} &  & 78.4 & 90.6 \\
PSP + EFC (Ours)  &  & 80.2 & 90.9 \\
PSP\_CRS \cite{zhao2017pyramid} &  \checkmark & 80.2 & 90.6 \\
PSP\_CRS + NetWarp \shortcite{gadde2017semantic}  & \checkmark & 80.5 & 91.0 \\ 
PSP\_CRS + GRFP \shortcite{nilsson2016semantic} & \checkmark & 80.6 & 90.8 \\ 
PSP\_CRS + EFC (Ours)  & \checkmark  & \textbf{81.0} & \textbf{91.2} \\
\hline
\hline
DeepLabv3+ \cite{chen2018encoder} & \checkmark &  82.1 & 92.0 \\
+ EFC (Ours) & \checkmark &  82.7 & 92.1 \\
+ VPLR \shortcite{zhu2019improving}& \checkmark &  \textbf{83.5} & \textbf{92.2} \\
\hline
\end{tabular}
\end{center}
\label{tab:cityscapes}
\end{table}

The ablation study results for segmentation are presented in Table~\ref{tab:ablation}. It can be seen that:
(1) The performance continuously increases when more components are used for video segmentation, showing the contribution of each part.
(2) Compared with the fixed FlowNetS, joint learning with the optical flow benefits the video segmentation, which shows the close relationship between these two tasks.
(3) The temporally consistent constraint has made huge improvements (a percentage of 1.3) to video segmentation, even without the use of occlusion mask.
(4) The improvements achieved by occlusion mask show that modeling of occlusion regions benefits the video segmentation.
(5) Both the use of more labeled data pairs and unlabeled data clearly lead to performance improvements, which provides evidence that our EFC model takes full advantage of video information.

For optical flow estimation, we make comparisons to five versions of our model:
(1) UL -- unsupervised learning of only the flow branch with the smooth loss and the photometric loss;
(2) UL + OE -- adding occlusion estimation ($O_\text{est}$) without the occlusion mask $O_{seg}$;
(3) UL + TC -- adding the segmentation branch and the temporal consistency module;
(4) UL + OE + TC -- our model without the occlusion handling module;
(5) EFC\_full -- our full model.

From Table~\ref{tab:kitti_ablation}, we can observe that: (1) Our model can learn in an unsupervised manner using only the optical flow branch.
(2) The segmentation branch and temporal consistent constraints greatly facilitate the learning of optical flow.
(3) A better occlusion estimation can further improve the performance of optical flow estimation.

\begin{figure}[t]
\begin{center}
\includegraphics[width=0.99\columnwidth]{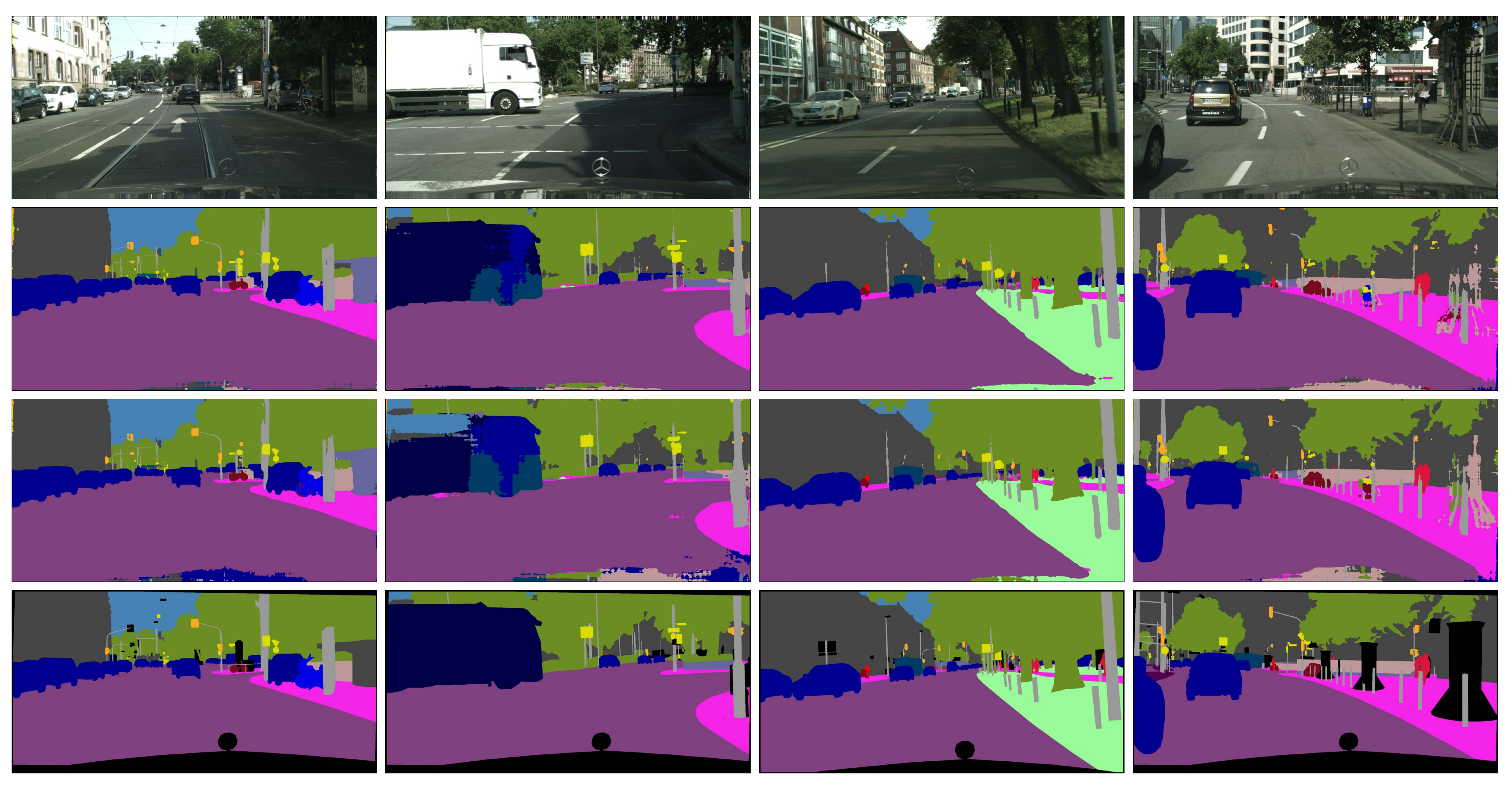}
\end{center}
\caption{Visual comparison on the Cityscapes validation set for segmentation. From top to bottom: original images, segmentation results of our model, PSPNet \cite{zhao2017pyramid} and the ground truth. Finely annotated data and single scale testing are used. Our approach yields large improvements in moving objects (motorcycle in the first column) and the category with less training data (truck in the second column).
}
\label{fig:segmentation}
\end{figure}

\subsection{Comparative Results}

\subsubsection{Video Semantic Segmentation}

We compare our video semantic segmentation model to the state-of-the-art alternatives on the challenging Cityscapes and CamVid datasets.

\noindent\textbf{Cityscapes} ~~ To validate the robustness of the proposed method on different network architectures, we used Dilation10 \cite{yu2015multi}, PSPNet \cite{zhao2017pyramid} and DeepLabv3+ \cite{chen2018encoder} as backbone network for the segmentation branch, respectively. In Table~\ref{tab:cityscapes} we show the quantitative comparison with a number of state-of-the-art video segmentation models.

We observe that: (1) With DeepLabv3+, PSPNet and Dilation10 as our backbones, our model are able to improve the mIoU score by 0.6, 1.8 and 2.1 respectively. Notice that our approach can be applied to any image semantic segmentation model for more accurate semantic segmentation. (2) VPLR \cite{zhu2019improving} first pre-trained on the Mapillary dataset, which contains
18,000 street-level scenes annotated images for autonomous driving. However, our model benefits from unlabeled data without the need of additional labeling costs. The performance can be further improved when we use coarsely annotated images. (3) Our segmentation model benefits from the spatial-temporal regularization in the feature space, thus there is no extra cost during the inference phase. All the other methods require additional modules and computational costs. Qualitative comparison is shown in Figure~\ref{fig:segmentation}.

\noindent\textbf{CamVid} ~~ We evaluate our method on the CamVid dataset and compare it with multiple video semantic segmentation methods. The comparative results are given in Table~\ref{tab:camvid}. Our model achieves the best result under the same setting.

\begin{table}[t]
\caption{Comparative results on the test set of CamVid for different video segmentation methods. All the methods are based on Dilation8 Network. Our model performs best and improve the mIoU score by 2.1 percentage.}
\begin{center}
\tabcolsep0.5cm
\begin{tabular}{c||c}
\hline Method & mIoU ($\%$) \\ \hline\hline
Dilation8 \cite{yu2015multi} & 65.3 \\
+ STFCN \shortcite{fayyaz2016stfcn} &  65.9 \\
+ GRFP \shortcite{nilsson2016semantic} &  66.1 \\
+ FSO \shortcite{kundu2016feature} &  66.1 \\
+ VPN \shortcite{jampani2017video} & 66.7 \\
+ NetWarp \shortcite{gadde2017semantic}   & 67.1 \\
\hline
+ EFC (ours) & \textbf{67.4} \\
\hline
\end{tabular}
\end{center}
\label{tab:camvid}
\end{table}

\begin{table}[t]
\caption{Average end-point error (EPE) on KITTI 2015 flow training set over non-occluded regions (Noc) and overall regions (All). Notation: `C' -- the FlyingChairs dataset, `S' -- the Sintel dataset, `T' -- the FlyingThings3D dataset, `K' -- the KITTI dataset, 'R' -- the RoamingImages dataset.}
\begin{center}
\tabcolsep0.3cm
\begin{tabular}{c||c|c|c}
\hline Method & Data & Noc & All \\ \hline\hline
EpicFlow \shortcite{revaud2015epicflow} & - & 4.45 & 9.57 \\
FlowNetS \shortcite{dosovitskiy2015flownet} & C+S & 8.12 & 14.19\\ %
FlowNet2 \shortcite{ilg2017flownet} & C+T & 4.93 & 10.06 \\
FlowNet2+ft \shortcite{ilg2017flownet} & C+T+K & - & 2.3 \\
PWC-Net \shortcite{sun2018pwc} & C+T & - & 10.35 \\
PWC-Net+ft \shortcite{sun2018pwc} & C+T+K & - & 2.16 \\

\hline
DSTFlow \shortcite{ren2017unsupervised} & K & 6.96 & 16.79 \\
GeoNet \shortcite{yin2018geonet} & K & 8.05 & 10.81 \\
OAULFlow \shortcite{wang2018occlusion} & K & - & 8.88 \\
Unflow \shortcite{meister2017unflow} & K & - & 8.80 \\
SC \shortcite{lai2019bridging} & K & 4.30 & 7.13 \\
EFC (ours)  & K & 3.93 & 7.05 \\
Back2Future \shortcite{janai2018unsupervised}  & R + K & 3.22 & 6.59 \\
SelFlow \cite{liu2019selflow} & S + K & -- & 4.84 \\

\hline
\end{tabular}
\end{center}
\label{tab:kitti}
\end{table}

\begin{figure}[t]
\centering
\includegraphics[width=0.99\columnwidth]{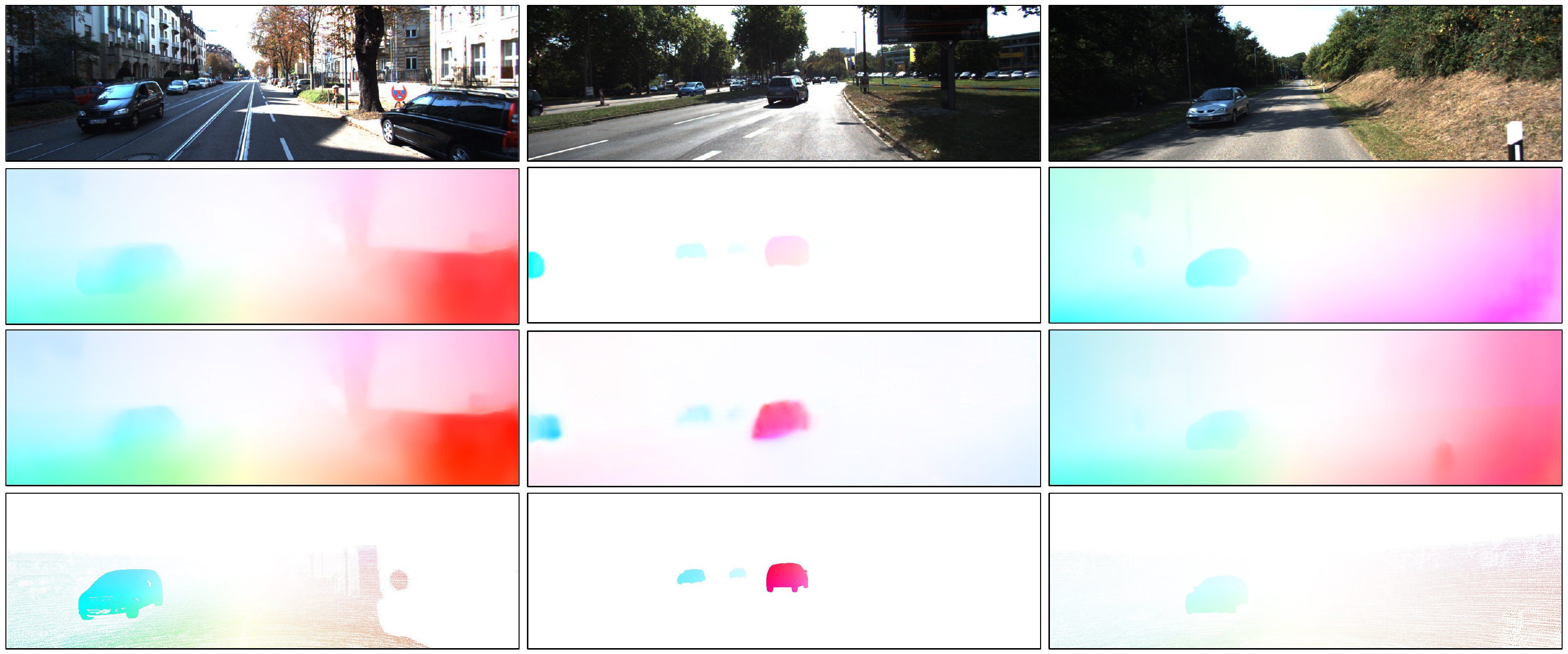}
\caption{Visual comparison on the KITTI dataset for optical flow. From top to bottom: original images, our results, GeoNet \cite{yin2018geonet} and ground truth. Our model estimate sharper motion boundaries than GeoNet. The middle column is an occluded case that the car is driving out of the camera scope, our model accurately handles the occlusion.}
\label{fig:flow}
\end{figure}

\subsubsection{Optical Flow Estimation}

To quantify how optical flow estimation benefits from the semantic segmentation, we evaluate the estimated flow on the KITTI dataset. Both supervised and unsupervised methods are included. As shown in Table~\ref{tab:kitti}, our model not only outperforms the existing unsupervised learning methods, but also yields comparable results with the Flownet2 \cite{ilg2017flownet} which is trained on FlyingChairs and FlyingThings3D datasets.
Following the common practice in \cite{ren2017unsupervised,yin2018geonet,wang2018occlusion,meister2017unflow,lai2019bridging}, we use no additional data and discard the whole sequence as long as it contains any test frames, while \cite{janai2018unsupervised,liu2019selflow} use the RoamingImages dataset and the Sintel dataset for pre-training, respectively. Besides, they use PWC-Net \cite{sun2018pwc} as the base model, which is powerful than FlowNetS.

The semantic segmentation brings semantic information to the optical flow estimation, which facilitates recovering sharp motion boundaries in the estimated flow. As shown in Figure~\ref{fig:flow}, our model fixes large regions of errors compared to \cite{yin2018geonet}.

\section{Conclusion}

In this paper, we propose a novel framework (EFC) for joint estimation of video semantic segmentation and optical flow. We observe that semantic segmentation introduces semantic information and helps model occlusion for more robust optical flow estimation. Meanwhile, non-occluded optical flow provides accurate pixel-level temporal correspondences to guarantee the temporal consistency of the segmentation. Moreover, we address the insufficient data utilization and the inefficiency issues through our framework. Extensive experiments have shown that our approach outperforms the state-of-the-art alternatives under the same settings in both tasks.

\section{Acknowledgements}

Zhiwu Lu is partially supported by National Natural Science Foundation of China (61976220, 61832017, and 61573363), and Beijing Outstanding Young Scientist Program (BJJWZYJH012019100020098). Ping Luo is partially supported by the HKU Seed Funding for Basic Research and SenseTime's Donation for Basic Research.

{\small
\bibliographystyle{aaai}
\bibliography{egbib}
}

\end{document}